\documentclass{article}
\usepackage[preprint]{iwbfstyle,amsmath,graphicx} 
\graphicspath{ {./images/} }

\title{Linear regression analysis of template aging in iris biometrics}
\name{Mateusz Trokielewicz$^{1,2}$}
\address{$^1$Biometrics Laboratory \\
Research and Academic Computer Network (NASK), Wawozowa 18, Warsaw, Poland\\
		$^2$Institute of Control and Computation Engineering\\
		Warsaw University of Technology, Nowowiejska 15/19, Warsaw, Poland}

\copyrightnotice{Manuscript accepted for publication at the IEEE IWBF 2015}

\begin{document}
%
\maketitle
\begin{abstract}
The aim of this work is to determine how vulnerable different iris coding methods are in relation to biometric template aging phenomenon. This is considered to be particularly important when the time lapse between gallery and probe samples extends significantly, to more than a few years.

Our experiments employ iris aging analysis conducted using three different iris recognition algorithms and a database of 583 samples from 58 irises collected up to nine years apart. To determine the degradation rates of similarity scores with extending time lapse and also in relation to multiple image quality and geometrical factors of sample images, a linear regression analysis was performed. 29 regression models have been tested with both the time parameter and geometrical factors being statistically significant in every model. Quality measures that showed statistically significant influence on the predicted variable were, depending on the method, image sharpness and local contrast or their mutual relations.

To our best knowledge, this is the first paper describing aging analysis using multiple regression models with data covering such a wide time period. Results presented suggest that template aging effect occurs in iris biometrics to a statistically significant extent. Image quality and geometrical factors may contribute to the degradation of similarity score. However, the estimate of time parameter showed statistical significance and similar value in each of the tested models. This reveals that the aging phenomenon may as well be unrelated to quality and geometrical measures of the image.

\end{abstract}
\section{Introduction}
\label{sec:intro}

Since the advent of iris recognition research there has been a profound belief in scientific community that selected features of the iris pattern are stable and not subject to changes over a person's lifetime. Safir and Flom mention this for the first time in their iris recognition patent dated on 1987 \cite{SafirFlom}. John Daugman presents a similar statement in his 1994 patent \cite{DaugmanPatent}. However, to this day no research has been presented that would prove these hypotheses.

Biometric algorithms reliability assessment is usually conducted using samples collected within a short time period (from days to months), but it is extremely difficult to evaluate performance of coding methods using images collected a few or more years apart. Given the long time lapse, it is particularly not easy to collect sufficient databases that span over many years and provide sample images from a large number of subjects.

In this paper we describe an evaluation of three iris coding methods in terms of their vulnerability to the iris template aging phenomenon, defined as a similarity score degradation with the extending time-lapse between gallery and probe images. We use 583 sample iris images from 58 irides collected up to nine years apart.

\section{Related work}
\label{sec:related}

Tome-Gonzalez \emph{et al.} \cite{gonzalez2008} evaluate the time impact on intra-class variability using dataset of 8128 iris images collected from 254 people over several periods covering one to four weeks. Authors report that for the Libor-Masek algorithm there is over 50\% increase in the False Rejection Rate (abbreviated FRR later on) caused by a visible shift in genuine inter-session similarity scores when compared to the intra-session.

Bowyer \emph{et al.} mention the term `template aging' in their paper concerning factors that contribute to degrading biometric recognition performance as one of `the accepted truths' about iris biometrics \cite{kevin2009factors}. However, they also point to the pupil dilation differences between images as a possible cause of the decrease in genuine similarity scores. Baker \emph{et al.} report that the average Hamming distances for iris template comparisons conducted 4 years after the gallery images were acquired are greater than when short time span is involved. This is true both when averaged for all irides and for each iris separately \cite{Baker2009}. Those studies consisted of 26 irides. The estimated FRR of the evaluated system rose about 75\% when compared to the scenario in which probe images were collected after a shorter (\emph{i.e.} a few days) time span. Researchers go further in their experiments, extending the database to 46 irides and dividing it into subsets containing data collected up to 120 days after the first enrollment and those collected after more than 1200 days \cite{Baker2013}. IrisBEE algorithm, Neurotechnology VeriEye and Cam-2 from 2006 ICE all showed a significant increase in FRR, while FAR remained mostly unaffected. VeriEye, being the most accurate method, noted an increase of about 70\%. Fenker \emph{et al.} broaden this research using dataset of 86 irides collected two years apart, reporting FNMR (False Non-Match Rate) for the VeriEye matcher 195\% to 457\% higher, depending on the acceptance threshold applied \cite{Fenker2011}. Different causes of such increase, including a visible dilating of the pupil with increased time lapse, are laid out. Finally, Fenker and Bowyer investigate a dataset that consists of images of 644 irides using a commercially available VeriEye method \cite{Fenker2012}. Authors create 4 sets of comparisons: between images collected no longer that several months apart, over 1 year apart and over 2 and 3 years apart, respectively. Each of them was then evaluated in reference to the set of comparisons between images collected with a short time span, noting increases in FRR as high as 50\%.

Czajka conducts an experiment using a dataset of 58 irides with time-span between acquisitions inside a class reaching over 8 years \cite{Czajka2013}. A proprietary BiomIrisSDK algorithm, VeriEye SDK and MIRLIN SDK are employed and degradation of the average genuine similarity score for all methods reaching as high as 45\%, is reported. In addition, Czajka creates a modified dataset with iris diameter unified throughout a given class, however, no statistically significant differences are found.

In contrast to most results stands the NIST IREX VI report \cite{IREX}, stating that 'recognition metrics are stable, consistent with the absence of widespread iris ageing' and that 'iris recognition of average individuals will remain viable over decades'. However, authors do not consider any factors, other that biological changes to the iris anatomy, as contributing to the aging phenomenon. Bowyer and Ortiz \cite{BowyerIREX} present a critique of this approach, pointing at several methodological errors in the regression analysis performed by the authors, as well as at it's non-compliance with operational practice.

Another possible factor that can have impact on iris recognition is the sensor ageing effect, discussed in \cite{sensor}. Authors claim that this has the potential to affect iris recognition due to the noise appearing in aged imaging sensors, however, no explicit trends or tendencies are found. 

Sazonova \emph{et al.} propose broadening aging studies by including quality measures calculated for iris images \cite{Sazonova}. Authors suggest that covariates like the number of occluded pixels, local contrast, average intensity of pixels and image sharpness may contribute to the degradation of recognition performance. To prove this hypothesis they conduct a linear regression analysis using two regression models: one containing only the time parameter, the latter being a combination of time and quality parameters. Researchers prove that there is a statistical significance of all regression parameters for at least one sample from a pair of images. This is true for Neurotechnology's VeriEye SDK and Masek's algorithm. Also, the time parameter estimate contributes to an average increase in genuine Hamming distance with the value of 0,0077 per year.

Being a rather novel approach to the topic of template aging, multimodal regression analysis aimed at finding as many contributing factors as possible certainly seems worth investigating. Our own work described in this paper takes inspiration from the experiments of Sazonova \emph{et al.} while taking it further with additional factors.

\section{Data}
\label{sec:data}

\begin{figure}[!t]
\centering
\includegraphics[width=0.48\textwidth]{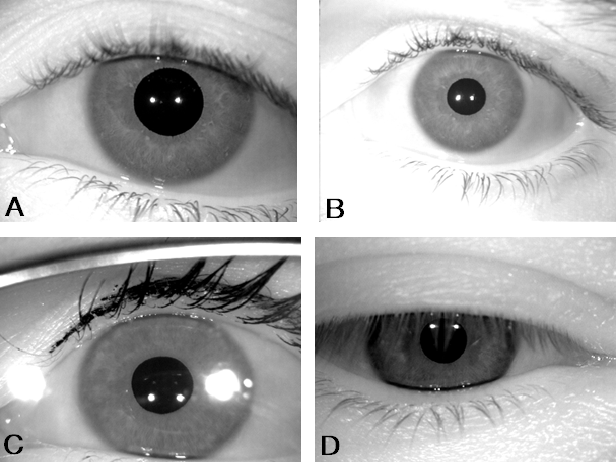}
\caption{Samples from sessions in 2003 (A) and 2010 (B). Sample images that were discarded from the dataset (C, D).}
\label{biobase}
\end{figure}

To conduct research described in this paper we use our own database consisting of iris images obtained from 35 persons (hence from 70 eyes). Depending on the class, there are 40 to 140 images of each iris and varying time span between image acquisitions that in some cases extends up to 2960 days. Majority of images have been collected in years 2003 -- 2004 and 2010 -- 2011 (see Fig. ~\ref{biobase}). This data have been carefully evaluated for poor quality samples not compliant with the ISO/IEC 19794-6:2011 \cite{ISO} and possibly clouding experimental results. We discarded all images out of focus, suffering from motion blur, with too little iris visible (due to closed eyelids or eyelash occlusion) and those with iris occluded by light reflections or eyeglasses.

Remaining dataset consisted of 583 images of 58 irides. Automatic segmentation employing BiomIrisSDK \cite{BiomIrisSDK} was performed on the data to locate the iris. However, in several cases iris has not been correctly localized. To make sure that in our analysis only aging-related factors contribute to the degradation of similarity scores (without additional dependencies such as failed segmentation), we performed a manual correction for those images, where automatic segmentation failed to produce an acceptable result.

\section{Experimental methodology}
\label{sec:methodology}

When performing an analysis of variance for the average similarity scores in selected data subsets, one needs to arbitrarily divide the dataset into several time periods. This approach does not include factors other than the time parameter. Also, there is little information about the extent of aging phenomenon outside the selected time periods, \emph{e.g.} how well will templates perform after another 4 or 8 years after the period covered by the database.

\subsection{Regression analysis predictors}

To determine the possible causes of similarity score decrease over time and to be able to predict the extent of these changes, we perform a linear regression analysis using several predictors that include the time parameter in combination with iris image quality and geometrical factors. As for quality factors we use the same as described in \cite{Sazonova}, namely: occlusion, local contrast, illumination and sharpness:

\begin{itemize}
\item \textbf{occlusion} is determined by the presence of eyelashes, eyelids and reflections that occlude the iris:
$OC = \frac{nm - N}{nm}$
where $n,m$ are polar image dimensions and N denotes the number of unoccluded iris pixels;
\item \textbf{local contrast}:
$LC = \sqrt{\frac{1}{N}\sum_{i=1}^{n}\sum_{j=1}^{m}\delta(I_{ij} - M_{ij})^2}$
where $\delta$ is the noise factor for pixel $i,j$ that becomes $0$ for the occluded pixel and $1$ otherwise, $I_{ij}$ is the pixel intensity and $n,m$ is the image size in pixels. $M_{ij}$ denotes the intensity median from 10-by-10 neighborhood centered in $i,j$;
\item \textbf{illumination} is to distinguish images being poorly illuminated and on that account potentially decreasing performance:
$IL = \frac{1}{N}\sum_{i=1}^{n}\sum_{j=1}^{m}I_{i,j}\delta$
where $I_{ij}$ and $\delta$ are defined as above;
\item \textbf{sharpness} to find images out of focus using the Laplacian of Gaussian:
$SH = \frac{1}{N}\sum_{i=1}^{n}\sum_{j=1}^{m}I^{LoG}_{ij}$
\end{itemize}
We also use two geometrical factors: iris and pupil radii (denoted as IR and PR, respectively) and their variability in a given image pair.


\subsection{Matchers}
In this study we use three different iris coding methods: BiomIrisSDK developed by Czajka \cite{BiomIrisSDK, CzajkaPacut}, Neurotechnology's VeriEye SDK \cite{VeriEye} and an implementation of Daugman's method \cite{Brize}. 

BiomIrisSDK employs Zak-Gabor wavelet packets to find (independently) the best space and frequency pairs when calculating the iris code. This methodology has also been employed to perform a manual correction in cases when automatic segmentation would yield poor results. Two 90-degree regions on the opposite sides of the iris are used for encoding. They were also modified in some cases to make sure that no occluded (non-iris) pixels are taken into account. Such attitude lets us hope that conclusions withdrawn from experiments will leverage time-related iris pattern alteration and not other factors, such as faulty segmentation. To our best knowledge, this is the first attempt of such approach. BiomIrisSDK produces results in a form of Hamming distance between compared iris codes, with values  near 0 for the same, and roughly 0.5 for different eyes. 

VeriEye SDK uses proprietary, unpublished encoding methodology with active shape based image segmentation. This matcher provides result in a form of similarity score, with values spreading from 0 (non-match in ideal case) to infinity (for two same images).

The third method is close to the original Daugman's idea based on filtering the image with 2D Gabor wavelets and creating the iris code using signs of real and imaginary parts of the filter response \cite{DaugmanArticle}. It employs real wavelets instead of complex ones, originally found in Daugman's work. It requires images in polar coordinates, therefore they have been prepared accordingly to localization results devised earlier with BiomIrisSDK segmentation and our corrections. As a consequence we ensure that the input data for this matcher represents only the proper regions of the iris.

\subsection{Linear regression modeling}
Due to the fact that not all of the examined matchers use an occlusion mask or take advantage of known localization results, different regression models are built for each method. 29 models are evaluated to propose a single and most accurate regression model for each method. For each of those models the $R^2$ statistic and significance levels for all regression factors are calculated (with $\alpha=0.05$).

For the Daugman's method 12 regression models are created. The first one ($D_{0}$) is identical to the model proposed by Sazonova in \cite{Sazonova}.
$$D_0 = \beta_0 + \beta_{1}t + \beta_{2}OC_{1} + \beta_{3}OC_{2} + \beta_{4}LC_{1} + \beta_{5}LC_{2} +$$
$$ + \beta_{6}IL_{1} + \beta_{7}IL_{2} + \beta_{8}SH_{1} + \beta_{8}SH_{2} + \epsilon$$
where $\beta_0$ is the intercept, $\beta_{i}$ are the regression coefficients for corresponding regression parameters, and $\epsilon$ stands for the noise. $D_{i}$ represents the predicted \emph{HD} value. For the sake of simplicity, all models are described in the same manner: $D_0$ through $D_{11}$, $B_0$ through $B_9$ and $V_0$ through $V_{11}$ for the Daugman, BiomIrisSDK and VeriEye SDK matchers, respectively.

The next step was to create four models ($D_1$ through $D_4$) and substitute sums of coefficients with multiplication (for $OC$) or differences ($LC$, $IL$ and $SH$), e.g.:
$$D_1 = \beta_0 + \beta_{1}t + \beta_{2}|OC_{1} * OC_{2}| + \beta_{3}LC_{1} + \beta_{4}LC_{2} +$$
$$+ \beta_{5}IL_{1} + \beta_{6}IL_{2} + \beta_{7}SH_{1} + \beta_{8}SH_{2} + \epsilon$$ 
This was then performed for the $LC$, $IL$ and $SH$ as well. In our opinion this approach seems more appropriate as the image order is not taken into account. It is also difficult to come up with any conclusion if one of the images in a pair yields statistical significance while the other does not. In model ($D_5$), this change is made for all parameters:
$$D_5 = \beta_0 + \beta_{1}t + \beta_{2}|OC_{1} * OC_{2}| + \beta_{3}|LC_{1} - LC_{2}| + $$
$$+ \beta_{4}|IL_{1} - IL_{2}| + \beta_{5}|SH_{1} + SH_{2}| + \epsilon$$  
Then a model incorporating only geometrical factors was built ($D_6$) and another one ($D_7$) being a combination of models $D_5$ and $D_6$ to evaluate a possibility of mutual impact of those two groups of factors (geometrical and quality).
$$D_6 = \beta_0 + \beta_{1}t + \beta_{2}|PR_{1} - PR_{2}| + \beta_{3}|IR_{1} - IR_{2}| + \epsilon$$
$$D_7 = \beta_0 + \beta_{1}t + \beta_{2}|OC_{1} * OC_{2}| + \beta_{3}|LC_{1} - LC_{2}| +$$
$$ + \beta_{4}|IL_{1} - IL_{2}| + \beta_{5}|SH_{1} + SH_{2}| + \beta_{6}|PR_{1} - PR_{2}| +$$
$$+ \beta_{7}|IR_{1} - IR_{2}| + \epsilon$$
Lastly, four additional models were created, each being the same as $D_7$, but deficient in one of the quality coefficients ($OC$, $LC$, $IL$ and $SH$). This is to determine whether such exclusion can increase significance of remaining factors.
$$D_8 = \beta_0 + \beta_{1}t + \beta_{2}|LC_{1} - LC_{2}| + \beta_{3}|IL_{1} - IL_{2}| + $$
$$+ \beta_{4}|SH_{1} + SH_{2}| + \beta_{5}|PR_{1} - PR_{2}| + \beta_{6}|IR_{1} - IR_{2}| + \epsilon$$
For the BiomIrisSDK similar models were created (yet without the $OC$ parameters, as this method does not employ polar masks) This also modifies calculations, as quality factors have to be computed using pixels of an entire image. The VeriEye matcher models were same as for the BiomIrisSDK, but geometrical factors were excluded from those models due to the unknown segmentation result. 


\section{Results}
\label{sec:results}
\subsection{Regression analysis results for selected models}
\subsubsection{Daugman's method}
For the sake of clarity only those models that yielded promising results are described below. In the original $D_0$ model for every pair of quality factors the regression coefficient one of them is not statistically significant (with \emph{p-value} over $0.23$) while the other is (\emph{p-value} close to zero). This is not true only for the $OC$ parameters, however, based on such results one can presume that this is not the most appropriate way to predict changes, as the regression parameters change when images in an image pair are being replaced with one another. In this model only the second image form a pair yields statistically significant regression parameters.

\begin{table*}
\scriptsize
\caption{$P$-values for $\beta_{i}$ estimates and $R^2$ statistics in selected regression models ('--' when parameter is not present).}
\label{daugman_results}
\vskip0.2cm
\centering
\begin{tabular}{|p{0.14\textwidth}|c|c|c|c|c|c|c|c|}
\hline
\textbf{Model number} & $t$ & $|OC_{1} * OC_{2}|$ & $|\Delta LC|$ & $|\Delta IL|$ & $|\Delta SH|$ & $|PR_{1} - PR_{2}|$ & $|IR_{1} - IR_{2}|$ & $R^2$ statistic\\
\hline
$D_{5}$ & 0.0000 & 0.0097 & 0.4805 & 0.2777 & 0.0042 & -- & -- & 0.216\\
\hline
$D_{6}$ & 0.0000 & -- & -- & -- & -- & 0.0000 & 0.0000& 0.218\\
\hline
$D_{7}$ & 0.0000 & 0.0565 & 0.4386 & 0.3613 & 0.0148 & 0.0000 & 0.0000 & 0.225\\
\hline
$D_{final}$ & 0.0000 & -- & -- & -- & 0.0000 & 0.0000 & 0.0000 & 0.225\\
\hline
$B_{5}$ & 0.0000 & -- & 0.0000 & 0.9368 & 0.0129 & -- & -- & 0.345\\
\hline
$B_{6}$ & 0.0000 & -- & -- & -- & -- & 0.0000 & 0.0000 & 0.293\\
\hline
$B_{final}$ & 0.0000 & -- & 0.0000 & -- & 0.0022 & 0.0006 & 0.0000 & 0.351\\
\hline
$V_{5}$ & 0.0000 & -- & 0.0000 & 0.7817 & 0.0000 & -- & -- & 0.352\\
\hline
$V_{9}$ & 0.0000 & -- & -- & 0.0011 & 0.0000 & -- & -- & 0.275\\
\hline
$V_{final}$ & 0.0000 & -- & 0.0000 & -- & -- & -- & -- & 0.352\\
\hline
\end{tabular}
\end{table*}

While experimenting with models $D_1$ through $D_4$, we found that replacing one of the pairs of factors with an absolute value of either a multiplication or a subtraction causes different effect, depending on the type of the factor: it can increase or decrease the statistical significance of this particular factor while increasing or decreasing significance of others. Due to the fact that there are possibly too many hidden relations between those coefficients, we came up with a conclusion that it would be best if all quality factors are represented in this way. This is done in model $D_5$. Here, the $|OC_1 * OC_2|$ and $|SH_1 - SH_2|$ factors gain the highest statistical significance, while the $|LC_1 - LC_2|$ and $|IL_1 - IL_2|$ factors are not statistically significant. We can thus assume that the former two should be included in the final regression model proposed for this coding methodology. 

The $D_6$ model incorporates only the time parameter and geometrical factors: pupil and iris radii, that being $|PR_1 - PR_2|$ and $|IR_1 - IR_2|$, respectively. There is a statistical significance for all of the parameters, which leads to the conclusion that those are also parameters that have to be put in a final regression model. 

Models $D_5$ and $D_6$ have been combined in model $D_7$ that proves further significance of the $OC$ and $SH$ factors while $LC$ and $IL$ remain statistically insignificant. Despite the fact the the $OC$ factor's \emph{p-value} is slightly higher than $\alpha=5\%$ it is later included in further models as we managed to show that the presence of certain factors might decrease the significance of others.

This becomes even more evident when evaluating models $D_8$ through $D_{11}$. When the LC parameter is removed, the $|OC_1 * OC_2|$ parameter's \emph{p-value} drops below $\alpha = 0.05$ again. Further investigation of these four models lets us observe that none of them incorporate $IL$ and $LC$ factors with statistical significance. Also, removing these parameters increases the significance of $OC$ and $SH$ parameters. 

To conclude, for the Daugman's coding methodology the $|OC_1 * OC_2|$ and $|SH_1 - SH_2|$ seem to be the best candidates for the final regression model in combination with both geometrical factors. The time parameter is statistically significant in every tested model ($p < 10^{-7}$), which proves its impact on the predicted Hamming distance between samples. With the estimate of $t = 0.000018$, this contributes to an increase in HD of 0.007 with each elapsed year.

\subsubsection{BiomIrisSDK}
In the first model for the BiomIrisSDK ($B_0$) both $LC$ factors gain statistical significance, unlike in all previous models. So are statistically significant the first parameters from the IL and SH parameter pairs, but not the second ones. This changes when instead of one $LC$ factor for each image a $|LC_1 - LC_2|$ factor is used. All of the model parameters are statistically significant then.

When introducing the $|OC_1 * OC_2|$ instead of two separate factors every regression parameter in a model showed statistical significance. However, replacing the image intensity in the same way did not produce any better results for the rest of the parameters. With model $B_5$ we find out that $LC$ and $SH$ factors expressed as $|LC_1 - LC_2|$ and $|SH_1 - SH_2|$ may be usable in the final model, as both of them represent high statistical significance.

Analogously to the Daugman's method, both geometrical factors (namely $|PR_1 - PR_2|$ and $|IR_1 - IR_2|$) show high statistical significance for the BiomIrisSDK matcher in every model. This suggests that these factors should be involved in building the final model as well.

\subsubsection{VeriEye SDK}
For the VeriEye matcher, the first model $V_0$ (the same as $D_0$, but without the $OC$ parameters) brings up only one factor that is statistically significant. That being said, we modified each of the models, replacing one pair of factors at a time as we did for the two former matchers. Model $V_2$ (with $|LC_1 - LC_2|$) yielded the best results with all regression parameters statistically significant ($p$-value $ < 0.05$).

In model $V_5$ all factors were replaced ($|LC_1 - LC_2|$, $|IL_1 - IL_2|$ and $|SH_1 - SH_2|$). The only quality factor that showed statistical significance was local contrast. This factor was the only one statistically significant in two more models, in which $|IL_1 - IL_2|$ and $|SH_1 - SH_2|$ were removed, one at a time. However, only model $V_9$ produced statistical significance for every regression parameter present in it and this model does not include $|LC_1 - LC_2|$ parameter. Yet because model $V_9$ presented lower $R^2$ statistic (approx. 0.352 versus 0.275) than models promoting the local contrast factor, we chose to include $LC$ in the final model built for the VeriEye matcher.

\subsection{Proposed final regression models}

After a detailed analysis of 29 regression models it is possible to put forward three with best results in terms of statistical significance of regression parameters. For the Daugman method we propose a following model:
$$D_{final} = \beta_0 + \beta_{1}t + \beta_{2}|SH_{1} - SH_{2}| + \beta_{3}|PR_{1} - PR_{2}| +$$
$$+ \beta_{4}|IR_{1} - IR_{2}| + \epsilon$$
It predicts the change in Hamming distance in relation to time (increase in HD of 0.007 each year), differences in image sharpness and differences in pupil and iris radii. All regression parameters are statistically significant.

As for the BiomIrisSDK, the $B_{final}$ model is considered to produce best results. It attempts to predict change in HD in terms of time (increase in HD of 0.007 each year), pupil and iris radii, local contrast and image sharpness. All factors are given in a form of a difference in covariates calculated for each image from a pair.
$$B_{final} = \beta_0 + \beta_{1}t + \beta_{2}|LC_{1} - LC_{2}| + \beta_{3}|SH_{1} - SH_{2}| +$$
$$+ \beta_{4}|PR_{1} - PR_{2}| + \beta_{5}|IR_{1} - IR_{2}| + \epsilon$$
Finally, model $V_{final}$ seems best for the VeriEye matcher. It incorporates only the time parameter (decrease in similarity score of about 11.3 each year) and a difference in local contrast of the two images. Those parameters show high statistical significance (with \emph{p-values} below machine accuracy). This model also yields the highest $R^2$ statistic of all three models proposed, being the most accurate in predicting the extent of the studied effect:
$$V_{final} = \beta_0 + \beta_{1}t + \beta_{2}|LC_{1} - LC_{2}| + \epsilon$$

\section{Discussion}
Results brought up in our work extend beyond most of the previous studies with a series of regression models for each of tested matchers, that attempt to to predict changes in similarity score as a function of time parameter in combination with geometrical and quality image factors. Each of the models was tested for the \emph{p-value} it produces for every regression parameter, to put forward a few models that give best results. 
		
Time parameter is statistically significant in every single model, which clearly expounds time impact on aging phenomenon, recognized as a perceivable degradation of genuine similarity score distributions. That being said, we found that aging might be autonomous from iris image quality and geometrical characteristics. Nevertheless, those components shall be taken into account throughout future studies since some combinations of them prove to be statistically significant in regression modeling.
		
Although some of the models produced statistical significance for all of their predictors, low $R^2$ statistics reveal that there may be miscellaneous factors that were not taken into account in our work or investigations of other researchers, but which may also contribute to the studied phenomenon. It is also critical to examine and expose mutual interrelations between applied regression parameters to come up with more sentience while constructing regression models.
		
Experimental evidence of iris template aging shall not be a starting point to diminish iris as a biometric characteristic. Future studies should focus on collecting vast databases including as many samples and embracing as long time periods as reasonably possible. Detailed research on the matter shall incorporate evaluating more coding methodologies and put forward attainable countermeasures to keep iris recognition a fast, reliable and secure biometric method.

\section{Acknowledgment}
This paper summarizes the author's B.Sc. project realized at Warsaw University of Technology in 2012 and 2013 and lead by Dr. Adam Czajka. The author would therefore like to thank Dr. Czajka cordially for his help, commitment and valuable insight during this project. This work is part of a wider iris aging research conducted at Warsaw University of Technology since 2003.

\bibliographystyle{IEEEbib}
\bibliography{refs}

\end{document}